\documentclass[final]{cvpr}

\usepackage{times}
\usepackage{epsfig}
\usepackage{graphicx}
\usepackage{amsmath}
\usepackage{amssymb}
\usepackage{multirow}
\usepackage{booktabs}
\usepackage[dvipsnames]{xcolor}

\usepackage[pagebackref=true,breaklinks=true,colorlinks,citecolor=RoyalBlue,bookmarks=false]{hyperref}

\DeclareMathOperator*{\argmin}{arg\,min}

\def\cvprPaperID{4936} % *** Enter the CVPR Paper ID here
\def\confYear{CVPR 2021}

\begin{document}
%%%%%%%%% TITLE
\title{Efficient Object Embedding for Spliced Image Retrieval} % Replace with your title

% \author[1]{Don Joe}
% \author[2]{Smith K.}
% \author[1]{Wanderer}
% \author[1]{Static}
% \affil[1]{TeX.SX}
% \affil[2]{Both on a bus}
% \date{}                     %% if you don't need date to appear
% \setcounter{Maxaffil}{0}
% \renewcommand\Affilfont{\itshape\small}
\author{Bor-Chun Chen$^{1,2}$ \qquad Zuxuan Wu$^{3}$\thanks{Work done during author was in Facebook AI.} \qquad Larry S. Davis$^1$ \qquad Ser-Nam Lim$^2$\\\\
$^1$University of Maryland, College Park \qquad $^2$Facebook AI \qquad $^3$Fudan University\\
{\tt\small \{sirius, lsd\}@cs.umd.edu, zxwu@fudan.edu.cn, sernamlim@fb.com}
}

\date{%
    $^1$Organization 1\\%
    $^2$Organization 2\\[2ex]%
    \today
}

\maketitle
\thispagestyle{empty}
\pagestyle{empty}

%%%%%%%%%% ABSTRACT
\begin{abstract}

Detecting spliced images is one of the emerging challenges in computer vision. Unlike prior methods that focus on detecting low-level artifacts generated during the manipulation process, we use an image retrieval approach to tackle this problem. When given a spliced query image, our goal is to retrieve the original image from a database of authentic images. To achieve this goal, we propose representing an image by its constituent objects based on the intuition that the finest granularity of manipulations is oftentimes at the object-level. We introduce a framework, object embeddings for spliced image retrieval (OE-SIR), that utilizes modern object detectors to localize object regions. Each region is then embedded and collectively used to represent the image. Further, we propose a student-teacher training paradigm for learning discriminative embeddings within object regions to avoid expensive multiple forward passes. Detailed analysis of the efficacy of different feature embedding models is also provided in this study. Extensive experimental results show that the OE-SIR achieves state-of-the-art performance in spliced image retrieval.

\end{abstract}

\section{Introduction}

With the proliferation of social media platforms and the availability of user-friendly image editing software, adversaries can now easily share spliced images on the Internet and reach millions of people with malicious intent to spread misinformation, disrupt democratic processes, and commit fraud. The ability to detect such spliced images is thus an increasingly important research area. Most existing work learns a mapping function between a spliced image and its corresponding label map, where each pixel in the map denotes whether the pixel has been modified or not~\cite{bappy2017exploiting,huh2018fighting,salloum2018image,zhou2018learning}. However, such training strategies require dense pixel-level annotations, which are expensive to obtain and thus prevent their abilities to scale. In this paper, we formulate splicing detection as an image retrieval task:
given a spliced query image and a large-scale image database, our goal is to retrieve images in the database that are authentic versions of the query image. We describe this as {\bf Spliced Image Retrieval (SIR)} problem. Once the original images are retrieved, we can then localize the spliced regions in these images by comparing the query and the retrieved images.

\begin{figure}[t]
\begin{center}
\includegraphics[width=0.9\linewidth]{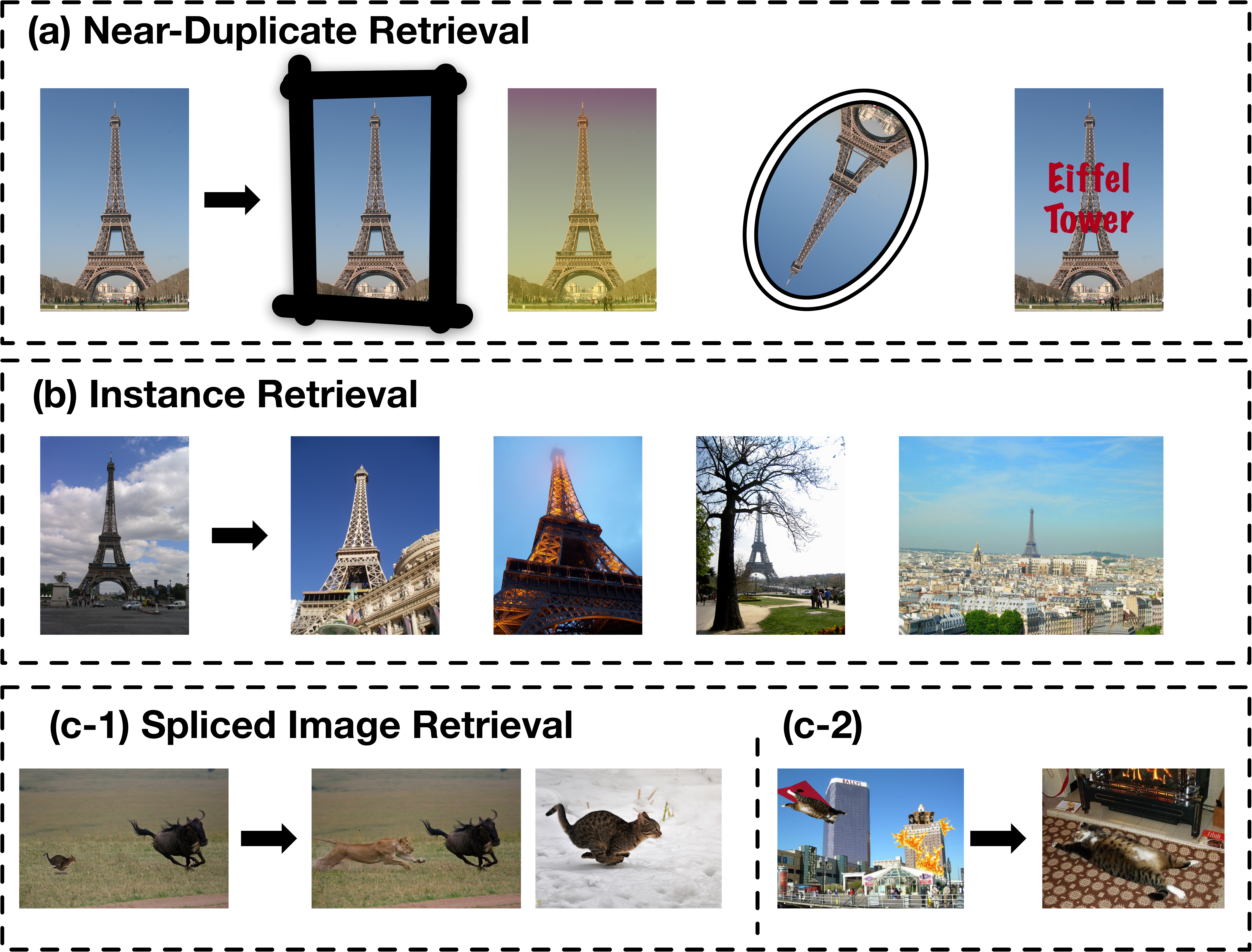}
\end{center}
\vspace{-0.1in}
  \caption{Three different types of image retrieval tasks. (a) The traditional image retrieval algorithm tries to retrieve near-duplicate images using low-level image statistics. (b) Instance retrieval tries to retrieve the same instance (e.g. building) under different viewpoint, illumination, and occlusion. (c) Spliced image retrieval (SIR) tries to retrieve authentic images used to create the spliced query image. Results from SIR contain images with large variation, so it is difficult to learn a single embedding that is suitable for the SIR task.}
\label{fig:1}
\end{figure}

In contrast to traditional image retrieval which usually focuses on retrieving near-duplicate images or images containing specific instances, SIR focuses on retrieving authentic images that were used to create the query. Figure~\ref{fig:1} shows examples of near-duplicate retrieval, instance retrieval, and SIR. As shown in the figure, SIR faces a different set of challenges compared to near duplicate or instance retrieval task. First, the query image may contain both manipulated as well as non-manipulated regions (cf. Figure~\ref{fig:1} (c-1)). When comparing with images in the database, the query image should be region-specific rather than using the entire image as in many current image retrieval systems~\cite{babenko2014neural,gordo2016deep,radenovic2016cnn,gordo2017end}.

Secondly, retrieved images may not overlap and query expansion~\cite{chum2007total}, a common practice in this research area, does not apply. Using Figure~\ref{fig:1} (c-1) as an example, if we were to use the first image to do query expansion, we might retrieve more images of similar lions and horses but we will not be able the retrieve the image of the cat running in the snow since two images do not share any overlapping content. Third, the query image and the authentic image might have extreme diverse backgrounds (cf. Figure~\ref{fig:1} (c-2)), which may cause traditional image retrieval algorithms to fail.

To mitigate these issues, we take advantage of recent advances in object detection \cite{dollar2014fast,ren2015faster,he2017mask} and propose Object Embeddings for Spliced Image Retrieval (OE-SIR). Instead of using a single global representation, OE-SIR generates object-level representations per object region that is then collectively used to represent the image. By representing an image at object-level granularity, and assuming that image manipulations are frequently done by manipulating objects (e.g., faces, logos, etc.), we can extract similar object embeddings from both spliced and authentic images, achieving the purpose of SIR.

Given detected objects, the next challenge lies in deriving robust feature representations for the task of spliced image retrieval. It is appealing to directly use features from detection networks as such embeddings are trained with additional location information and it is computationally efficient with a single forward pass. However, while extensive studies have been conducted on image retrieval, most of them were focused only on embeddings provided by classification networks \cite{babenko2014neural,arandjelovic2016netvlad,tolias2015particular,babenko2015aggregating,gordo2016deep,yue2015exploiting}. In light of this, we provide a detailed analysis of embeddings derived from different pre-trained object detectors, and compare them with image classification models. Our analysis shows that even though the object detection networks are trained with additional annotations, the resulting embeddings are significantly worse than those from classification models for image retrieval. This suggests a computationally expensive two-step process for SIR---detecting objects with object detectors and then encode them with pretrained classification models.

We propose a student-teacher training regime to explore the best of both worlds for computational efficiency, \ie, reliable bounding boxes produced by detectors and discriminative features computed with classification models. This is achieved by training a lightweight student network on top of the detection model that projects feature maps of the detection model into a more discriminative feature space guided by the teacher model. The student network decouples feature learning from localization, preserving the discriminative power of the features for classification.

The contributions of this work include: (1) We introduce the task of spliced image retrieval and propose OE-SIR that derives object-level embeddings. (2) We provide a detailed analysis of embeddings extracted from different pre-trained models and show that embeddings extracted from object detection models are less discriminative than those from image classification models. (3) We show that OE-SIR can outperform traditional image retrieval baseline and achieve significantly better results with two SIR datasets. (4) OE-SIR demonstrates state-of-the-art performance in detecting spliced regions by utilizing the original image.

\section{Background and Related Work}
\noindent {\bf Image forensics.}
Finding manipulated images is an important topic in media forensics research. Traditional approaches \cite{ye2007detecting,mahdian2009using,ferrara2012image} usually focus on finding low-level artifacts in the manipulated images. Recently, with the success of deep learning in computer vision, many people also turn to deep learning algorithms \cite{zhou2018learning,huh2018fighting,bappy2017exploiting,salloum2018image,wu2018busternet} to detect manipulated images. Specifically, Zhou \etal \cite{zhou2018learning} utilize object detection framework \cite{ren2015faster} to detect manipulated region in images. In contrast to their approach, we use an object detection framework for learning object embeddings and use the embeddings to retrieve spliced images. Most previous approaches try to detect manipulation from a single image. There are a few recent studies focusing on provenance analysis \cite{moreira2018image}, constructing a graphical relationship of all manipulated images. However, these approaches usually treat detecting manipulated content as a segmentation task, which requires dense annotations as supervision. In this paper, we formulate the problem as a retrieval task without the need to use pixel-level annotations.

\vspace{0.05in}
\noindent {\bf Content-based image retrieval.} Image retrieval aims to identify relevant images from an image database given a query image based on the image content. Early work \cite{rui1999image} used global color and texture statistics such as color histogram and Gabor wavelet transform to represent the image. Later advances on instance retrieval using local features \cite{lowe2004distinctive} and indexing methods \cite{sivic2003video,jegou2010improving,jegou2011product} achieved robustness against illumination and geometric variations. With the recent broad adoption of convolutional neural networks (CNN), different techniques have been proposed for global feature extraction \cite{babenko2014neural,babenko2015aggregating,tolias2015particular,arandjelovic2016netvlad,gordo2016deep,radenovic2016cnn,gordo2017end}, local feature extraction \cite{noh2017large,mishchuk2017working,zhang2017learning}, embedding learning \cite{movshovitz2017no,wu2017sampling,wang2017deep,ge2018deep}, as well as geometric alignment \cite{rocco2017convolutional,rocco2018end,melekhov2018dgc} using deep networks. Zheng \etal \cite{zheng2018sift} provide a comprehensive review of recent approaches towards image retrieval. Different from traditional image retrieval using either global features or local features, our approach generates a few discriminative object embeddings utilizing object detection models and it is aiming for SIR.

\vspace{0.05in}
\noindent {\bf Representation learning from large-scale datasets.} Previous works mainly studied the transferability of embeddings extracted from classification models that have been trained on datasets such as ImageNet to other tasks \cite{donahue2014decaf,sharif2014cnn,yosinski2014transferable,huh2016makes,azizpour2016factors,kornblith2018better}. For instance, \cite{sharif2014cnn} reports comprehensive results of applying embeddings from the ImageNet-trained classification model to object detection, scene recognition, and image retrieval. In contrast, the efficacy of embeddings obtained from object detection models trained on large-scale datasets such as COCO \cite{lin2014microsoft} and OpenImages \cite{OpenImages} has not been widely studied. In this work, we provide an analysis of embeddings extracted from different models pre-trained on large-scale datasets for the retrieval task.

\vspace{0.05in}
\noindent {\bf Object detection} aims to detect different objects in an input image. Girshick \etal \cite{girshick2014rich} proposed one of the first deep learning based object detection models, R-CNN, which improved the accuracy significantly compared to traditional methods \cite{dalal2005histograms,felzenszwalb2009object,dollar2014fast}. Since then many enhancements \cite{ren2015faster,redmon2016you,lin2017feature,singh2018sniper} have been made to improve the accuracy as well as the training/inference time. Teichmann \etal \cite{teichmann2018detect} utilized a specialized landmark detection model to aggregate deep local features \cite{noh2017large} for landmark retrieval. A comprehensive survey of recent deep learning based object detection methods can be found in \cite{liu2018deep}. By taking advantage of recent success in object detection, our model can learn discriminative object-level embeddings for image retrieval. Joint detection and feature extraction has recently been used for person search tasks \cite{chen2020norm, dong2020bi}. However, these approaches requires annotations of bounding boxes as well as fine-grained person identities in the boxes. Therefore, these approaches can not directly apply to our task.

\vspace{0.05in}
\noindent {\bf Knowledge distillation} \cite{bucilua2006model,ba2014deep,hinton2015distilling,romero2014fitnets,chen2017learning} compress a complex model into a simpler one while maintaining the accuracy of the model. Bucilua \etal \cite{bucilua2006model} first proposed to train a single model to mimic the outputs of an ensemble of models. Ba \etal \cite{ba2014deep} adopted a similar idea to compress deep neural networks. Hinton \etal \cite{hinton2015distilling} further generalized the idea with temperature cross-entropy loss. Our student-teacher approach is related to knowledge distillation, which learns a simple student model to mimic the output of a complex one. What is different is that we leverage a detection network to provide additional guidance during training, which we show is effective for training the student network.

\begin{figure}[t]
\begin{center}
\includegraphics[width=1.0\linewidth]{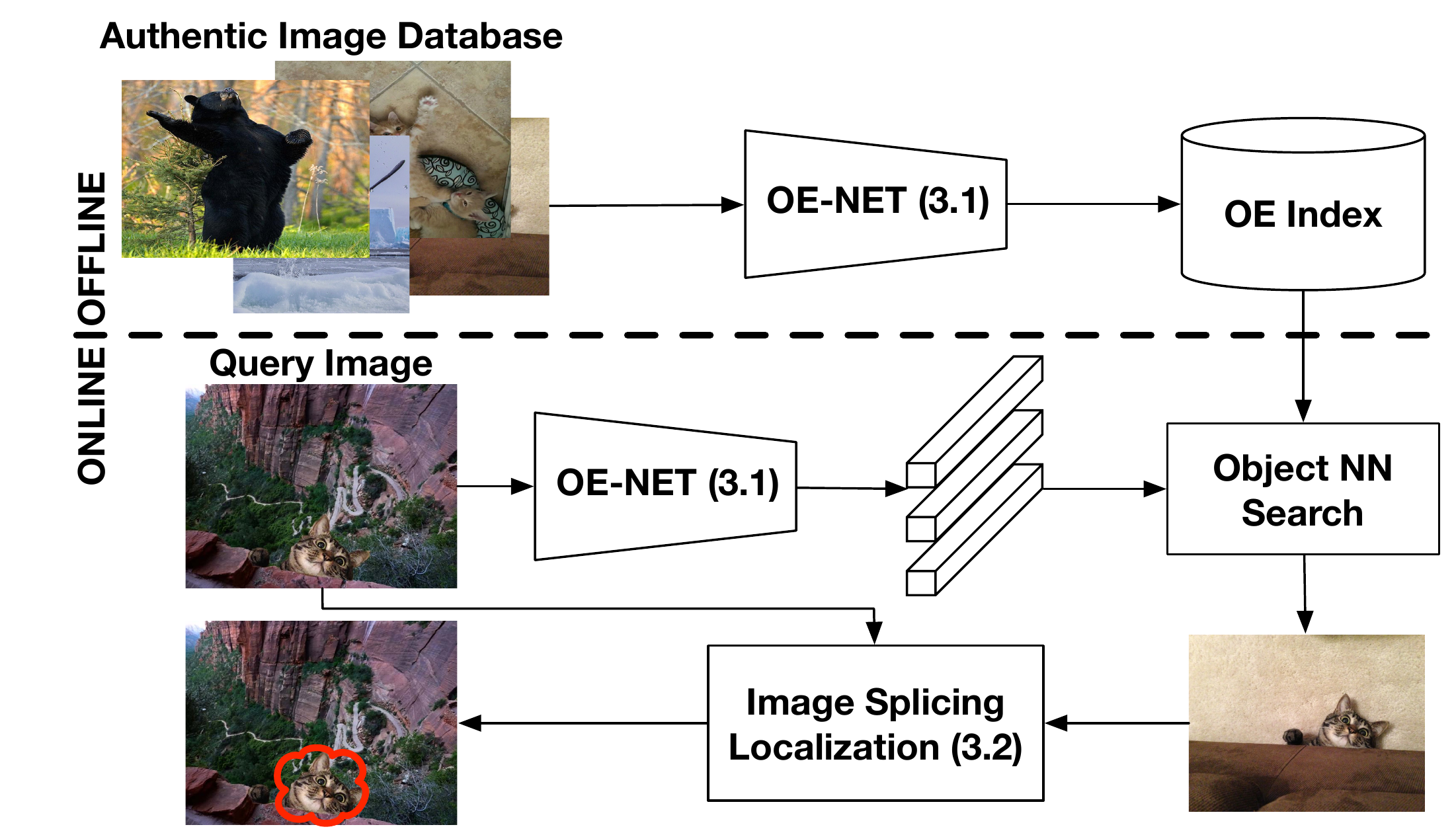}
\end{center}
   \caption{Overview of the proposed OE-SIR framework. A set of authentic images go through the proposed object embedding network (OE-NET) to build an object index offline. When a query image comes in, it first goes through the same network to extract object embeddings, and these embeddings are used to retrieve the authentic image with the object-level nearest neighbor search. Once the original image is retrieved, we can compare it with the query images to localize the spliced region.}
\label{fig:system}
\end{figure}

\section{Method}
\label{sec:method}
Given an image, our goal is to learn a feature embedding which models the image at the object-level such that it can be used to detect whether an object is spliced.
Figure~\ref{fig:system} shows the overview of our proposed OE-SIR framework. First, an object index is built with all available authentic images using the object embedding network described in Section~\ref{sec:model}. When a query image arrives,
an object embedding is computed with the same network, and then used to perform an object-level nearest neighbor search to retrieve authentic images. Finally, by comparing the query image with retrieved authentic images, we can localize the spliced region as described in Section~\ref{sec:retrieval}.

\subsection{Object Embedding Network (OE-NET)}
{\bf Object Detection and Feature Extraction Model.}
\label{sec:model}
The first step of OE-SIR is to train an object detection model $M_o$ and a feature extraction model $M_f$:
\begin{align}
    B, S = M_o(I), & & C = M_f(I),
\end{align}
where $B \in R^{n \times 4}$ denotes the bounding box coordinates for $n$ predicted objects in an image $I$, $S \in N^{n}$ is the object index, and $C \in R^{w \times h \times d}$ is a convolution feature map. In addition, $w,h,d$ is the width, height and the number of channels of the feature map.
We adopt the Faster-RCNN \cite{ren2015faster} object detection framework by minimizing the following multi-task loss during training for $M_o$:
\begin{align}
L(\{p_i\}, \{ti\}) = &\frac{1}{N_{cls}}\sum_i{L_{cls}(p_i, p_i^*)} + & \\  & \lambda\frac{1}{N_{reg}}\sum_ip_i^*{L_{reg}(t_i, t_i^*)},
\end{align}
where $L_{reg}$ is the bounding box regression loss and $L_{cls}$ is the classification loss, $p_i$, $p_i^*$ are the predict class label and ground truth label; $t_i$, $t_i^*$ are the predict box label and ground truth. The loss is minimized with SGD on standard detection datasets.

For $M_f$, since we do not have additional training data available, we utilize a pre-trained image classification model (\ie, ResNet \cite{he2016deep}) as our feature extraction network. We provide a detailed analysis of how we select our feature extraction model in Section~\ref{sec:analysis}. After both detection and feature extraction model are trained, we can extract object-level embeddings using the ROIAlign layer \cite{he2017mask} with $B$ as hard attention over the feature map $C$:
\begin{align}
X = ROIAlign(C, B),
\end{align}
where $X \in R^{(n \times d)}$ are object embeddings of the image, which contain $n$ predicted objects.

While it is possible to use a shared model for both object detection and feature extraction, we find that training two separate models provides many benefits. First, as we show in Section~\ref{sec:analysis}, jointly learning classification and localization reduces the discriminative power of the embeddings. Therefore, separate models ensure that we have better embeddings for retrieval. Second, since the detection model and the feature extraction model are independent, we can change the feature extraction model for a different task without retraining the object detection model. However, despite these advantages, such a two-step process is computationally expensive, requiring two forward passes for the same image. This limits the deployment of such models in resource-constrained environments such as mobile devices, robots, \etc.

\begin{figure}[t]
\begin{center}
\includegraphics[width=0.9\linewidth]{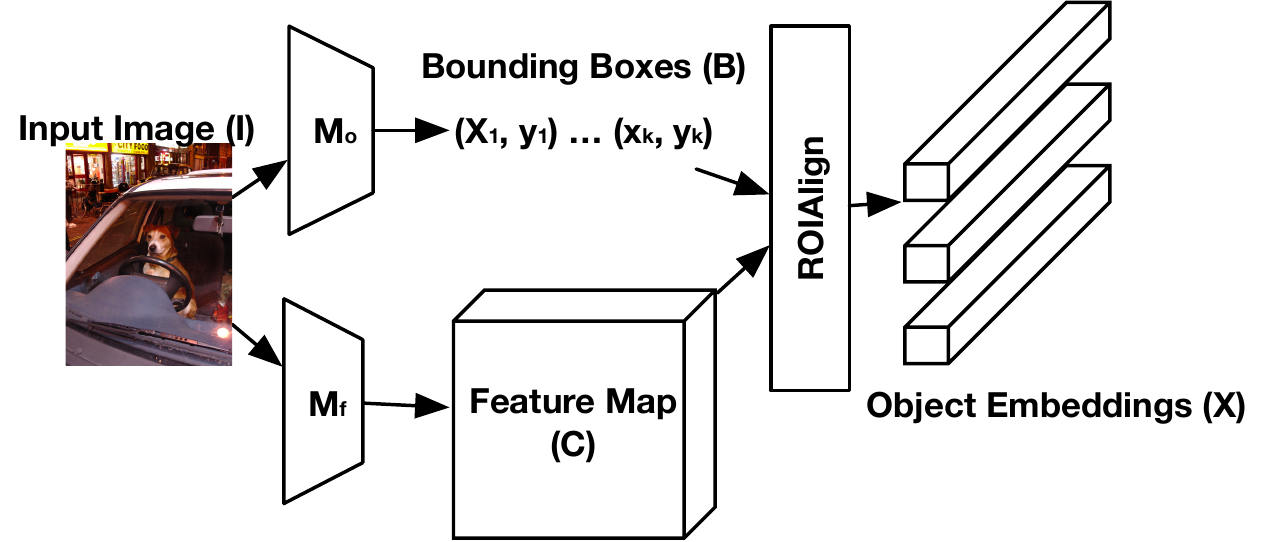}
\end{center}
   \caption{Configuration of the proposed object embedding network. The image first goes through object detector ($M_o$) to extract object bounding boxes, and a separate feature network ($M_f$) to extract discriminative feature map. We then use the detected bounding boxes to extract object embeddings from the feature map with ROIAlign layer.}
\label{fig:oen}
\end{figure}

\begin{figure}[t]
\begin{center}
\includegraphics[width=.9\linewidth]{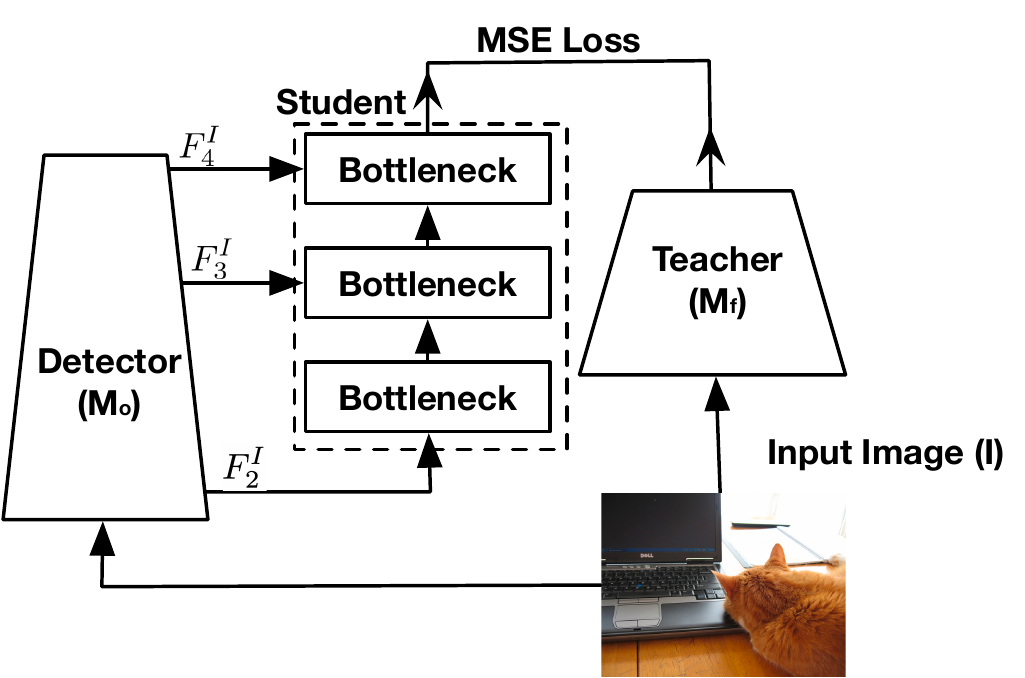}
\end{center}
   \caption{Knowledge distillation with the student network. Lightweight student network ($f$) utilize the information from detector backbone to learn discriminative feature map from the teacher network with mean square loss. Once the student model is trained, we can use single forward pass to extract discriminative object embeddings.}
\label{fig:student}
\end{figure}

{\bf Efficient Object Embeddings Extraction.}
We now introduce how to use a single model which explores the best of both worlds---robust bounding box detection with object detectors and discriminative feature computation with a classification model---such that given an image, object-embeddings can be efficiently computed with a single forward pass. Towards this goal, we use knowledge distillation \cite{hinton2015distilling} to save computation. One straightforward way is to train a student network completely from scratch to mimic the outputs of the classification model. However, this defeats the purpose of using a single model and neglects useful information from the backbone of the detector. Instead, we introduce a guided framework by using features from the detection backbone to train a student model.

More specifically,  we consider the classification model as a teacher and we attach a lightweight branch to the detector as a student model to approximate outputs from the teacher model. The reasons for using a parallel branch as the student network are two-fold: (1) we wish to reuse information from detector backbones to guide the training of the student model; (2) the student network can produce discriminative features for retrieval task while preserving the ability of the main detector branch to generate accurate bounding boxes. 

Formally, given an image $I$, the student network $f(F_2^I, F_3^I, F_4^I; \theta_s)$ is a lightweight model parameterized by the weight $\theta_s$. It takes as inputs feature maps after the second residual stage of the detector model $F_i^I$ to approximate the outputs of the classification model $M_f(I)$ with three bottleneck convolutional layers. We denote the outputs of the $i$-th ($i\in \{1, 2, 3\}$) bottleneck-layer as $y_i$.
To effectively leverage information from the multi-scale feature maps of the detector model, before feeding $y_i$ to the next layer, we modify it in an addition manner:
\begin{align}
    y_i = y_i + F^I_{i+2}, ~~~\text{if}~  1 \leq i \leq 2,
\end{align}
where $F^I_{i+2}$ denotes the outputs of the feature maps of the 3-rd ($i=1$) and 4-th ($i=2$) residual stage of the detector. Here, we assume the guidance has the same dimension as the layer output of the student model. For different dimensions, a linear transformation is applied to map them into the same space. Finally, the outputs of the student network are used to approximate the classification model by minimizing the following mean-squared loss:
\begin{equation}
    \min_{\theta_s} \sum_{I} || f(F_2^I, F_3^I, F_4^I; \theta_s) - M_f(I)||_2.
\end{equation}
Since the student model is optimized with multi-scale guidance from the detector, it can utilize both high-level and low-level features to learn discriminative features efficiently.

\subsection{Spliced Image Retrieval and Localization}
\label{sec:retrieval}
Given a query image represented by $n$ object embeddings $I_q = \{x_1^q, x_2^q, ..., x_n^q\}$, and a database image $I_d = \{x_1^d, x_2^d, ..., x_m^d\}$, the distance between the image pair is calculated as the minimum distance between pairwise object embeddings:
\begin{align}
    D(I_q, I_d) = \min_{i, j}{||x_i^{q} - x_j^d||_2^2}.
\label{eq:distance}
\end{align}
Note that while we are using multiple embeddings per image, $m$ and $n$ are usually really small and we can use quantization and indexing \cite{sablayrolles2018spreading} to speed up the retrieval process. For application with 100 million images and 32 bytes per embeddings with average eight embeddings per image, the total storage requirement is only around 24G, which can be easily stored in the memory of a single server.

\vspace{0.05in}
\noindent {\bf Localization.} We use Eq.~\ref{eq:distance} to retrieve the top-1 database image, $I_d^*$, which gives us a matching object pair:
\begin{align}
    (i, j) = \argmin_{i, j}{||x_i^q - x_j^d||_2^2}.
\label{eq:localization}
\end{align}
We can then use the bounding box $(b_i, b_j)$ information to estimate the geometric transformation $T$ between the image pair \cite{philbin2007object}. The final localization map $Q$ can then be generated by comparing convolutional feature map:
\begin{align}
    Q = ||M_f(T(I_q)) - M_f(I_d^*)||_2^2.
\label{eq:manipulated}
\end{align}
After comparison, we apply a threshold to Q to generate a binary mask for localization.

\section{Experimental Results}
\subsection{Feature Extraction Model}
\label{sec:analysis}

Harvesting data and annotations for splicing detection is expensive, and thus OE-SIR is built upon \emph{pre-trained} models that are widely available. Unlike specific tasks such as landmark retrieval~\cite{radenovic2018revisiting}, where there are usually additional sets of landmark images to train a good model, it is imperative that we select a good feature extraction model that provides discriminative embeddings. Here, we first provide a detailed analysis of embeddings extracted from different pre-trained models trained on image classification, object detection, and instance segmentation models using four common image retrieval benchmarks.

\vspace{0.04in}
\noindent {\bf Retrieval benchmark.} We consider four datasets for benchmarking, including USCB bird dataset \cite{WelinderEtal2010} ({\bf CUB200}), Stanford car dataset \cite{KrauseStarkDengFei-Fei_3DRR2013} ({\bf Cars196}), and two landmark datasets, $\mathcal{R}$Oxford5K \cite{radenovic2018revisiting} ({\bf $\mathcal{R}$Oxf}) and $\mathcal{R}$Pairs6K \cite{radenovic2018revisiting} ({\bf $\mathcal{R}$Par}).
For CUB200 and Cars196, we follow the same protocol in \cite{oh2016deep} and use leave-one-out partitions to evaluate every image in the test set. For $\mathcal{R}$Oxford5K and $\mathcal{R}$Paris6K, we follow the medium protocol described in \cite{radenovic2018revisiting}, using 70 and 55 images as queries, 4,993 and 6,322 images as database.  We use mean average precision (mAP) to measure the performance of different embeddings.

\vspace{0.04in}
\noindent {\bf Pre-trained models.} We consider seven different pre-trained models including (1) {\bf Faster-RCNN} \cite{ren2015faster} and (2) Faster-RCNN with feature pyramid networks \cite{lin2017feature} ({\bf Faster-RCNN-FPN}) trained on COCO \cite{lin2014microsoft}, (3) {\bf Mask-RCNN} \cite{he2017mask} and (4) Mask-RCNN with feature pyramid networks ({\bf Mask-RCNN-FPN}) trained on COCO with bounding box and mask annotations, and (5) {\bf ResNet50} \cite{he2016deep} trained on ImageNet. To control the effect of different training data, we also compare with (6) {\bf Faster-RCNN} and (7) {\bf ResNet50} trained with the same dataset (OpenImagesV4 \cite{OpenImages}).
We adopt open source implementation\footnote{https://github.com/facebookresearch/detectron2} of Faster-RCNN and Mask-RCNN with ResNet50 as a backbone feature extractor for all our detection and segmentation models. For all Faster-RCNN and Mask-RCNN models, we use weights from the ImageNet classification model to initialize the backbone network and use the default 3$\times$ learning rate schedule to train the models. We use images from OpenImagesV4 to learn a PCA whitening matrix for post processing \cite{jegou2012negative}.

\begin{table}[t]
  \begin{center}
  \resizebox{\columnwidth}{!}{ \begin{tabular}{lcccccc}
    \toprule
    \multirow{2}{*}{Model} & Train Set & \multirow{2}{*}{$\mathcal{R}$Oxf} & \multirow{2}{*}{$\mathcal{R}$Par} & \multirow{2}{*}{CUB200} & \multirow{2}{*}{Cars196}\\
     &(\# of Img. / Cls.)&&&&\\
    \midrule
    {Faster-RCNN \cite{ren2015faster}} &    &18.7&    28.3&    4.1&    2.4\\
    Faster-RCNN-FPN \cite{lin2017feature} & COCO  &20.4&31.0&    3.3&    3.1\\    
    Mask-RCNN \cite{he2017mask}       & (330K / 80) &20.7&    33.0&3.0&2.4\\
    Mask-RCNN-FPN \cite{lin2017feature}& &34.2&    48.1&    2.9&    3.6\\    
    \multirow{2}{*}{ResNet50 \cite{he2016deep}} & ImageNet   &\multirow{2}{*}{\bf 40.1}    & \multirow{2}{*}{\bf 57.3}    &\multirow{2}{*}{\bf 21.2}    &\multirow{2}{*}{\bf 11.1}\\
    & (1.2M / 1K) &&&&\\
    \midrule
    Faster-RCNN \cite{ren2015faster}  & OpenImagesV4  &19.5    & 32.3    &4.7&2.2    \\
    {ResNet50 \cite{he2016deep}} & (1.7M / 601) & {\bf 41.2} & {\bf 61.2} & {\bf 19.3} & {\bf11.0}\\ 
    \bottomrule
  \end{tabular}}
    \end{center}
    \vspace{-0.1in}
  \caption{Image retrieval performance (mAP) with embeddings extracted from different pre-trained models for four different retrieval benchmarks. Even though all detection and instance segmentation models are initialized with weights trained on ImageNet classification dataset, the embeddings learned from these models perform significantly worse than embeddings learned from the classification model.}  
  \label{tab:analysis}
\end{table}

During test time, we first resized the image to a maximum size of $1024 \times 1024$, followed by extracting features from conv5\_3 layer \cite{he2016deep} and using max-pooling to produce image embeddings from different pre-trained models. We then use cosine similarity between embeddings for retrieval ranking. We do not apply any post-processing tricks such as multi-scale ensemble and query expansion except PCA whitening.

\vspace{0.04in}
\noindent {\bf Embeddings comparison.} Table~\ref{tab:analysis} shows the mean average precision of different models when used as feature extractors on the four retrieval benchmarks. Comparing Faster-RCNN (COCO) and Mask-RCNN (COCO), we note that additional mask annotations decrease the performance of the embeddings on some of the dataset, suggesting that localization constraints could hurt the retrieval performance further. Also, by increasing the size of the training set from COCO to OpenImagesV4, the Faster-RCNN performance improves on some datasets but degrades on other datasets. Most importantly, although all the models are initialized with weights trained on ImageNet classification, embeddings extracted from detection and segmentation models perform significantly worse than the embeddings from the ImageNet classification model. Even when comparing Faster-RCNN (OpenImages) with ResNet50 (OpenImages) which are trained with the same training data, but with Faster-RCNN utilizing more human annotations (\ie, bounding boxes), embeddings learned from the classification model still significantly outperform embeddings learned from the detection model. This suggests that enforcing both classification and localization during training compromises the discriminative ability of the embedding. Consequently, decoupling localization and classification might be crucial for learning embeddings that are effective for image retrieval as we mentioned in Section~\ref{sec:model}. Based on the analysis, we select the ResNet50 classification model as our feature extraction model for SIR.

Note that different spatial pooling techniques \cite{radenovic2018revisiting} and post-processing steps such as dimensionality reduction \cite{jegou2012negative} have been shown to greatly affect retrieval performance. Furthermore, embeddings from different layers of the network also perform differently. We provide detailed analysis in the supplementary material for selecting these parameters.

\subsection{Student Networks}
\begin{table}
  
    \begin{center}
  \resizebox{\columnwidth}{!}{ \begin{tabular}{ccccc}
    \toprule
     & FLOPs & \# Params. & {$\mathcal{R}$Oxf} & {$\mathcal{R}$Par}\\
    \midrule
        Faster-RCNN  & - & - & 25.4  &34.4 \\
        $S_1$   &$1.49 \times 10^9$ & $8.02 \times 10^6$ & 32.1   &55.2\\
        $S_2$  &$1.13 \times 10^9$ & $7.93 \times 10^6$ & 43.3    &56.3\\
        $S_3$  &$1.13 \times 10^9$ & $7.93 \times 10^6$ & {\bf 50.2} & {\bf 65.2} \\
        \midrule
        Teacher (ResNet50) & $3.33\times 10^9$ & $8.54 \times 10^6$ & 53.4 & 69.7\\
    \bottomrule
  \end{tabular}}
    \end{center}
    \vspace{-0.1in}
  \caption{FLOPs, number of parameters and mAP for different student models. The performance of the proposed $S_3$ achieves better performance while using fewer FLOPs and model parameters comparing to two other baseline student model.}  
  \label{tab:student}
  \vspace{-10pt}
\end{table}

We compare the student network proposed in Section~\ref{sec:model} with two baseline versions: (1) $S_1$: Lightweight network with five bottleneck layers without any guidance information from the detector backbone. (2) $S_2$: Lightweight network with three bottleneck layers with taken $F_2^I$ from the detector backbone as an input feature map. (3) $S_3$: proposed network with multi-scale inputs from the detector backbone. See supplementary material for an illustration of these networks. We use images from the OpenImageV4 dataset to train different student models. Note that the training of the student model is unsupervised and does not require any manual annotations. We use Adam \cite{kingma2014adam} optimizer with a learning rate of 1e-3 and batch size of 64 to train all the student models for 20,000 iterations. During inference, we use the minimum distance between pairwise object embedding derived from the student networks to retrieve database images. Table~\ref{tab:student} shows the performance of different student models in terms of mAP as well as the computational cost and model parameters evaluating on the landmark dataset {$\mathcal{R}$Oxf} and {$\mathcal{R}$Par}. $S_1$ achieves the worst performance and it struggles to learn discriminative embeddings. $S_2$ achieves slightly better performance than $S_1$ by reusing the low-level feature maps from the detector. Utilizing the guidance from multi-scale feature maps of the detection model, our best student model is $S_3$, which achieves up to 93.5\% of the original performance, but only requires one-third of the FLOPs used by the teacher networks. Note that mAP of the teacher model is higher than the image-level retrieval results in Section~\ref{sec:analysis} which demonstrates the importance of utilizing object embeddings. Additional results on landmark retrieval can be found in the supplementary material.

\subsection{Spliced Image Retrieval}

To demonstrate the effectiveness of our approach for SIR, we conduct experiments on two different benchmarks. (1) {\bf COCO-Fake}. COCO-Fake consists of 58 query images with spliced objects generated by the method described in \cite{chen2019toward}, and 10,000 authentic images from the COCO dataset, including images that are used to create the queries. (2) Photoshop Image Retrieval dataset ({\bf PIR}). The images are collected from the publicly available PS-Battles Dataset \cite{heller2018psBattles}. We use 70,389 spliced images as queries and 10,592 authentic images as the database. Since we mostly care about whether we can retrieve the correct match in the top rank, we use recall at K (R@K) as our evaluation metric, which shows the percentage of queries that have the correct match in the top K rank. Note that since only one image is expected to be retrieved per query, recall at K metric is identical to accuracy at K.

\begin{figure*}[t]
\begin{center}
\includegraphics[width=0.86\linewidth]{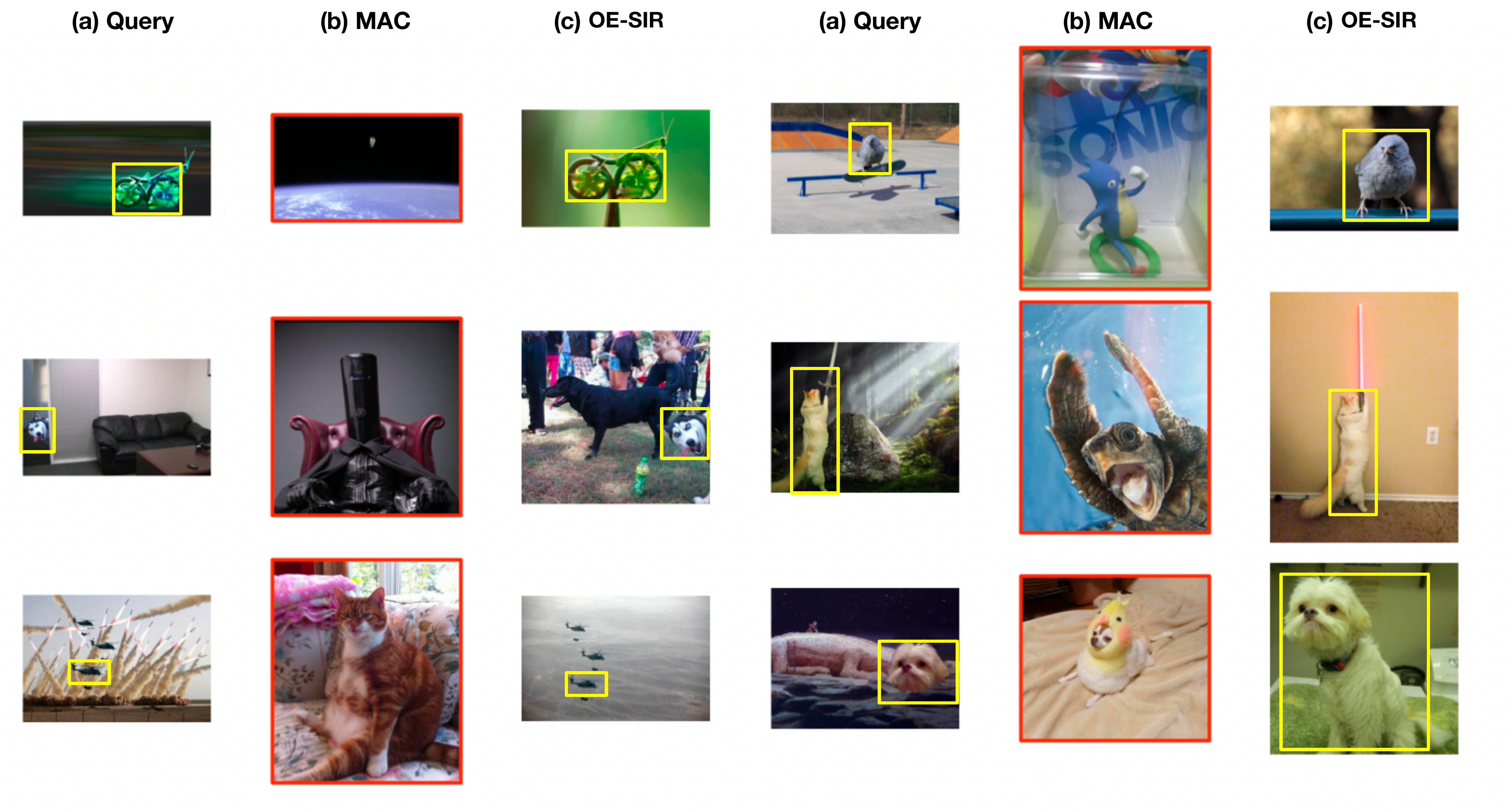}
\end{center}
\vspace{-0.1in}
   \caption{Example images and rank one results in the PIR dataset. (a) spliced query images. (b) Rank-1 result from MAC (c) Rank-1 result from OE-SIR. The red border indicates incorrect matches and the yellow bounding box shows the matching objects. OE-SIR can retrieve the original images while global image embeddings fail to retrieve the correct match due to background clutter.}
\label{fig:dataset}
\end{figure*}

Table~\ref{tab:fake} shows retrieval results compared to different image retrieval methods using the same ImageNet feature extraction model including (1) SPoC descriptors \cite{babenko2015aggregating}, (2) maximum activations of convolutions (MAC) \cite{razavian2016visual}, (3) regional maximum activation of convolutions (R-MAC), and (4) generalized mean pooling (GeM) \cite{radenovic2018fine}. On COCO-Fake, our model performs significantly better because all query images are with small spliced objects and the traditional image retrieval approach fails in this case. On PIR, where it contains in-the-wild spliced images, our approach still achieves better performance. Figure~\ref{fig:dataset} shows some examples of the retrieval result. Figure~\ref{fig:dataset} (a) are the query images; Figure~\ref{fig:dataset} (b) show rank-1 retrieved results by MAC. MAC retrieves images with similar scenes but fails to retrieve original images that contain the spliced objects from the query image. Figure~\ref{fig:dataset} (c) shows the rank-1 results retrieved by OE-SIR. More qualitative results including some failure cases can be found in the supplementary material. Note that while most of the examples in the dataset are spliced for entertainment purposes, spliced images can easily be used for malicious intent to spread misinformation.

\vspace{0.04in}
\noindent {\bf Object-level retrieval.} We also compare with two additional baseline methods that utilize the same object-level retrieval framework as the proposed method: (1) OE-HoG: After object detection, we extract histogram of oriented gradients \cite{dalal2005histograms} from each object to build object index and use object-level HoG for retrieval. (2) OE-FasterRNN: Directly using object features from the Faster-RCNN network for object-level retrieval. By utilizing the object-level search, simple handcraft features with object embeddings (OE-HoG) can achieve competitive performance compare to deep learning based image retrieval approach, which demonstrate the importance of the object-level search. On the other hand, OE-FasterRCNN performs worse than the proposed method, which also confirms the finding in Section~\ref{sec:analysis} that jointly learning classification and localization degrades the discriminative power of the embeddings.

\begin{table}[t]
  \begin{center}
  \begin{tabular}{lcccccc}
    \toprule
     \multirow{2}{*} {Method} &  \multicolumn{2}{c}{COCO-Fake} & \multicolumn{2}{c}{PIR}\\
     &  R@1 & R@10 &  R@1 & R@10\\
    \midrule
    SPoC \cite{babenko2015aggregating}    & 29.3 & 34.3 & 43.2 & 46.6 \\
    MAC \cite{razavian2016visual} &  29.3 & 34.8 & 52.6 & 59.9\\
    R-MAC \cite{tolias2015particular}   & 37.9 & 42.5 & 51.6 & 58.5 \\
    GeM \cite{radenovic2018fine}  &  37.9 & 43.7 & 48.2 & 54.2 \\
    \midrule
    OE-HoG \cite{dalal2005histograms} & 43.1 & 48.3 & 49.8 & 53.6\\
    OE-FasterRCNN \cite{ren2015faster} & 39.7 & 55.1 & 48.7 & 54.8 \\
    OE-SIR (Ours) & {\bf 70.7} & {\bf 84.5} & {\bf 58.6} & {\bf 67.7}\\
    \bottomrule
  \end{tabular}
    \end{center}
    \vspace{-0.1in}
  \caption{Performance on COCO-Fake and PIR dataset. Our approach outperforms other baseline approaches for retrieving authentic images with spliced objects.} 
  \label{tab:fake}
\end{table}

\begin{table}[t]
  
  \begin{center}
  \begin{tabular}{cccc}
    \toprule
    \multirow{2}{*} {\# of objects} & \multicolumn{3}{c}{PIR}\\
       &  R@1 & R@10 & R@100\\
    \midrule
    1 & 54.7&62.7&69.5\\
    2 &56.1&64.6&71.1\\
    4 & 57.6&66.3&72.9\\
    8 & {\bf 58.6} & 67.7 & 74.1\\
    16 & 58.4 & {\bf 67.8} & {\bf 74.7}\\
    \bottomrule
  \end{tabular}
  \end{center}
  \vspace{-0.1in}
  \caption{Performance of OE-SIR with different numbers of object embeddings by varying the detection threshold. The model with more embeddings achieves higher performance while requiring more memory storage.}  
  \label{tab:fake_num_obj}
\end{table}

\vspace{0.04in}
\noindent {\bf Number of object embeddings.} Table~\ref{tab:fake_num_obj} shows the performance of OE-SIR when using different numbers of object embeddings. We select up to $k$ objects in each image based on the confidence score of the detection model. Using more embeddings results in a higher recall, however, it also requires more memory. We found that using up to 8 object embeddings per image is a good trade-off since it requires a reasonable amount of memory and increasing the number of embeddings provides little performance gain.

\begin{figure}[t]
\begin{center}
\includegraphics[width=0.8\linewidth]{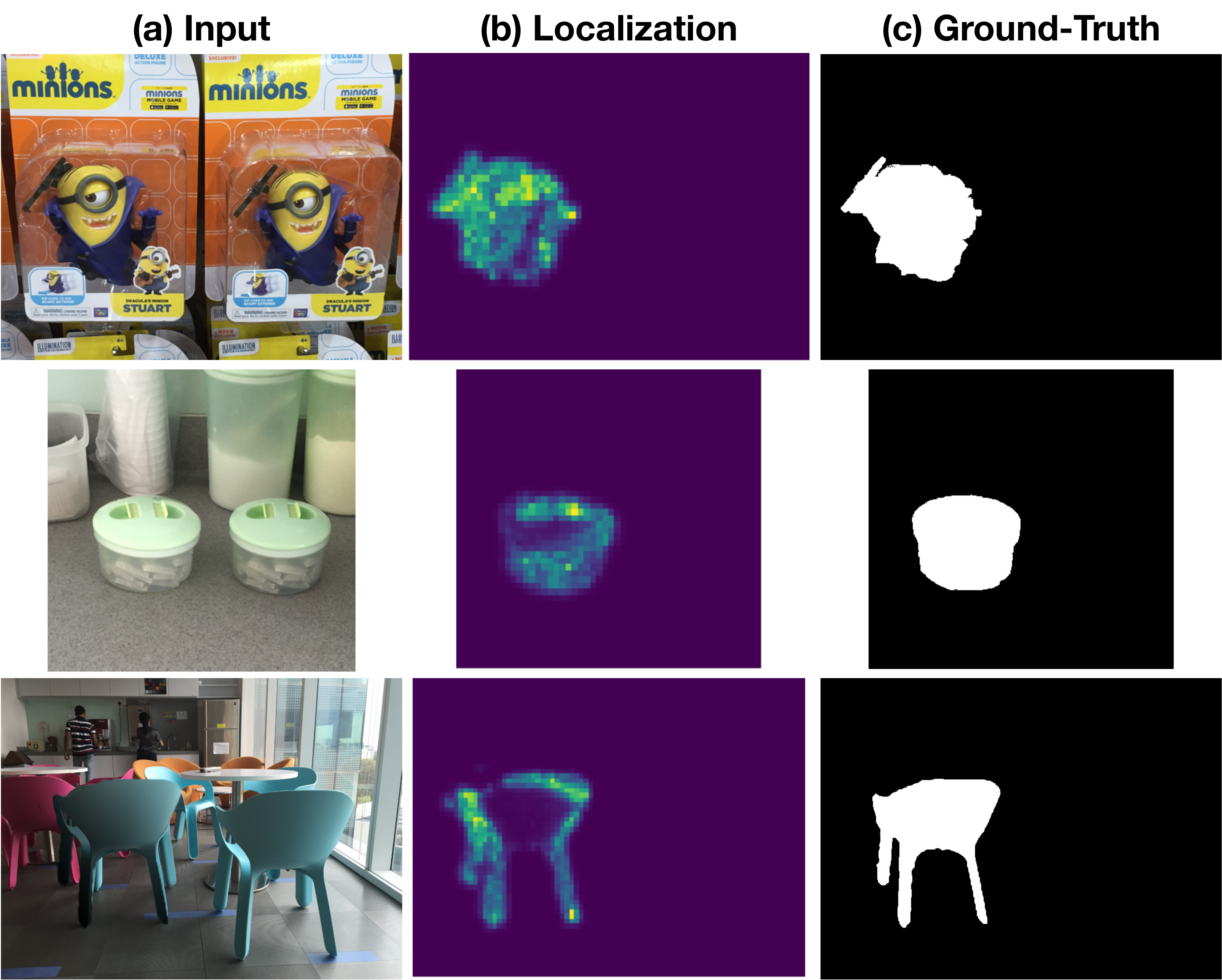}
\end{center}
\vspace{-0.1in}
   \caption{Qualitative results of image splicing localization on the COVERAGE dataset. The proposed method is able to correctly localize the spliced region by comparing the spliced image with the retrieved original image.}
   \vspace{-10pt}

\label{fig:localization}
\end{figure}

\begin{table}
\begin{center}
\begin{tabular}{ccc}
\toprule
Method & MCC & F1\\
\midrule
NOI \cite{mahdian2009using} & 0.172 & 0.269   \\
CFA \cite{ferrara2012image} & 0.050 & 0.190  \\
RGB-N \cite{zhou2018learning} & 0.334 & 0.379 \\
Self-Consistency \cite{huh2018fighting} & 0.102 & 0.276  \\
GSR-Net \cite{zhou2018generate}  & 0.439 & 0.489 \\
\midrule
OE-SIR (Ours) & {\bf 0.721} & {\bf 0.732} \\
\bottomrule

\hline
\end{tabular}
\end{center}
\vspace{-0.1in}
\caption{Image splicing localization performance. The proposed method significantly outperforms other state-of-the-art methods because our model can utilize the retrieved original image to localize the spliced regions. Note that our goal is to show the ability of image splicing localization using the retrieval approach, and our number is not directly comparable to previous methods since we utilize the original image.}
\vspace{-10pt}
\label{tab:localization}
\end{table}

\subsection{Image Splicing Localization}
\label{sec:localization}
\vspace{0.04in}
\noindent {\bf Dataset.} We show the performance of image splicing localization in a widely used image forensics dataset, COVERAGE \cite{wen2016coverage} dataset. COVERAGE contains 100 spliced images with its original version generate by copy-move manipulation. The spliced objects are used to superimpose similar objects in the original authentic images and thus are challenging for humans to recognize visually without close inspection. Figure~\ref{fig:localization} (a) shows some example of images in the COVERAGE dataset.

\vspace{0.04in}
\noindent {\bf Evaluation metrics.} We use the pixel-level F1 score and MCC as the evaluation metrics when comparing to other approaches and we follow the same measurement as \cite{zhou2018generate}, by varying the prediction threshold to get a binary prediction mask and report the optimal score over the whole dataset.

\vspace{0.04in}
\noindent {\bf Comparisons to state-of-the-art methods.} We compare several state-of-the-art image manipulation detection algorithms including (1) NoI \cite{mahdian2009using}: A noise inconsistency method which predicts regions as manipulated where the local noise is inconsistent with authentic regions. (2) CFA \cite{ferrara2012image}: A CFA based method that estimates the internal CFA pattern of the camera for every patch in the image and segments out the regions with anomalous CFA features as manipulated regions. (3) RBG-N \cite{zhou2018learning}: A two-stream Faster R-CNN based approach which combines features from the RGB and noise channel to make the final prediction. (4) Self-Consistency \cite{huh2018fighting}: A self-consistency approach that utilizes metadata to learn features useful for manipulation localization. The prediction is made patch by patch and post-processing like mean-shift \cite{cheng1995mean} is used to obtain the pixel-level manipulation prediction. (5) GSR-NET~\cite{zhou2018generate}: a segmentation-based approach with the generative adversarial network. (6) OE-SIR: Our spliced image retrieval approach. We first retrieve from the database (cf. Eq.~\ref{eq:distance}) the top-1 database image. Following Eq.~\ref{eq:localization} and \ref{eq:manipulated}, we then obtain the binary localization map corresponding to the top-1 image, after which we compute the MCC and F1 score. Table~\ref{tab:localization} shows the performance of different methods. Our approach outperforms other state-of-the-art methods by a significant margin because we can effectively utilize a database of original images. As we can see in Figure~\ref{fig:localization} (a) many of these spliced images are quite challenging even for a human. However, it becomes easier if we can compare it with the original images. Figure~\ref{fig:localization} (b) shows some qualitative results of the proposed method.

\section{Conclusion}
We describe an alternative approach to spliced image detection by casting it as an image retrieval task. Extensive experiments show the effectiveness of the proposed approach compares to the baseline algorithm. In addition, this paper provides a comprehensive analysis of features from pre-train models, allowing for a better selection of embedding models, which could be useful even for other tasks. Future research direction includes analysis of the effect of different object detectors as well as utilizing unsupervised or self-supervised methods to learn better features for the OE-SIR framework.

{\small
\bibliographystyle{ieee_fullname}
\bibliography{cvpr}

\begin{thebibliography}{10}\itemsep=-1pt

\bibitem{arandjelovic2016netvlad}
Relja Arandjelovic, Petr Gronat, Akihiko Torii, Tomas Pajdla, and Josef Sivic.
\newblock Netvlad: Cnn architecture for weakly supervised place recognition.
\newblock In {\em Proceedings of the IEEE Conference on Computer Vision and
  Pattern Recognition}, pages 5297--5307, 2016.

\bibitem{azizpour2016factors}
Hossein Azizpour, Ali~Sharif Razavian, Josephine Sullivan, Atsuto Maki, and
  Stefan Carlsson.
\newblock Factors of transferability for a generic convnet representation.
\newblock {\em IEEE transactions on pattern analysis and machine intelligence},
  38(9):1790--1802, 2016.

\bibitem{ba2014deep}
Jimmy Ba and Rich Caruana.
\newblock Do deep nets really need to be deep?
\newblock In {\em Advances in neural information processing systems}, pages
  2654--2662, 2014.

\bibitem{babenko2015aggregating}
Artem Babenko and Victor Lempitsky.
\newblock Aggregating local deep features for image retrieval.
\newblock In {\em Proceedings of the IEEE international conference on computer
  vision}, pages 1269--1277, 2015.

\bibitem{babenko2014neural}
Artem Babenko, Anton Slesarev, Alexandr Chigorin, and Victor Lempitsky.
\newblock Neural codes for image retrieval.
\newblock In {\em European conference on computer vision}, pages 584--599.
  Springer, 2014.

\bibitem{bappy2017exploiting}
Jawadul~H Bappy, Amit~K Roy-Chowdhury, Jason Bunk, Lakshmanan Nataraj, and BS
  Manjunath.
\newblock Exploiting spatial structure for localizing manipulated image
  regions.
\newblock In {\em Proceedings of the IEEE international conference on computer
  vision}, pages 4970--4979, 2017.

\bibitem{bucilua2006model}
Cristian Buciluǎ, Rich Caruana, and Alexandru Niculescu-Mizil.
\newblock Model compression.
\newblock In {\em Proceedings of the 12th ACM SIGKDD international conference
  on Knowledge discovery and data mining}, pages 535--541. ACM, 2006.

\bibitem{chen2019toward}
Bor-Chun Chen and Andrew Kae.
\newblock Toward realistic image compositing with adversarial learning.
\newblock In {\em Proceedings of the IEEE Conference on Computer Vision and
  Pattern Recognition}, pages 8415--8424, 2019.

\bibitem{chen2020norm}
Di Chen, Shanshan Zhang, Jian Yang, and Bernt Schiele.
\newblock Norm-aware embedding for efficient person search.
\newblock In {\em Proceedings of the IEEE/CVF Conference on Computer Vision and
  Pattern Recognition}, pages 12615--12624, 2020.

\bibitem{chen2017learning}
Guobin Chen, Wongun Choi, Xiang Yu, Tony Han, and Manmohan Chandraker.
\newblock Learning efficient object detection models with knowledge
  distillation.
\newblock In {\em Advances in Neural Information Processing Systems}, pages
  742--751, 2017.

\bibitem{cheng1995mean}
Yizong Cheng.
\newblock Mean shift, mode seeking, and clustering.
\newblock {\em IEEE transactions on pattern analysis and machine intelligence},
  17(8):790--799, 1995.

\bibitem{chum2007total}
Ondrej Chum, James Philbin, Josef Sivic, Michael Isard, and Andrew Zisserman.
\newblock Total recall: Automatic query expansion with a generative feature
  model for object retrieval.
\newblock In {\em 2007 IEEE 11th International Conference on Computer Vision},
  pages 1--8. IEEE, 2007.

\bibitem{dalal2005histograms}
Navneet Dalal and Bill Triggs.
\newblock Histograms of oriented gradients for human detection.
\newblock In {\em international Conference on computer vision \& Pattern
  Recognition (CVPR'05)}, volume~1, pages 886--893. IEEE Computer Society,
  2005.

\bibitem{dollar2014fast}
Piotr Doll{\'a}r, Ron Appel, Serge Belongie, and Pietro Perona.
\newblock Fast feature pyramids for object detection.
\newblock {\em IEEE transactions on pattern analysis and machine intelligence},
  36(8):1532--1545, 2014.

\bibitem{donahue2014decaf}
Jeff Donahue, Yangqing Jia, Oriol Vinyals, Judy Hoffman, Ning Zhang, Eric
  Tzeng, and Trevor Darrell.
\newblock Decaf: A deep convolutional activation feature for generic visual
  recognition.
\newblock In {\em International conference on machine learning}, pages
  647--655, 2014.

\bibitem{dong2020bi}
Wenkai Dong, Zhaoxiang Zhang, Chunfeng Song, and Tieniu Tan.
\newblock Bi-directional interaction network for person search.
\newblock In {\em Proceedings of the IEEE/CVF Conference on Computer Vision and
  Pattern Recognition}, pages 2839--2848, 2020.

\bibitem{felzenszwalb2009object}
Pedro~F Felzenszwalb, Ross~B Girshick, David McAllester, and Deva Ramanan.
\newblock Object detection with discriminatively trained part-based models.
\newblock {\em IEEE transactions on pattern analysis and machine intelligence},
  32(9):1627--1645, 2009.

\bibitem{ferrara2012image}
Pasquale Ferrara, Tiziano Bianchi, Alessia De~Rosa, and Alessandro Piva.
\newblock Image forgery localization via fine-grained analysis of cfa
  artifacts.
\newblock {\em IEEE Transactions on Information Forensics and Security},
  7(5):1566--1577, 2012.

\bibitem{ge2018deep}
Weifeng Ge.
\newblock Deep metric learning with hierarchical triplet loss.
\newblock In {\em Proceedings of the European Conference on Computer Vision
  (ECCV)}, pages 269--285, 2018.

\bibitem{girshick2014rich}
Ross Girshick, Jeff Donahue, Trevor Darrell, and Jitendra Malik.
\newblock Rich feature hierarchies for accurate object detection and semantic
  segmentation.
\newblock In {\em Proceedings of the IEEE conference on computer vision and
  pattern recognition}, pages 580--587, 2014.

\bibitem{gordo2016deep}
Albert Gordo, Jon Almaz{\'a}n, Jerome Revaud, and Diane Larlus.
\newblock Deep image retrieval: Learning global representations for image
  search.
\newblock In {\em European conference on computer vision}, pages 241--257.
  Springer, 2016.

\bibitem{gordo2017end}
Albert Gordo, Jon Almazan, Jerome Revaud, and Diane Larlus.
\newblock End-to-end learning of deep visual representations for image
  retrieval.
\newblock {\em International Journal of Computer Vision}, 124(2):237--254,
  2017.

\bibitem{he2017mask}
Kaiming He, Georgia Gkioxari, Piotr Doll{\'a}r, and Ross Girshick.
\newblock Mask r-cnn.
\newblock In {\em Proceedings of the IEEE international conference on computer
  vision}, pages 2961--2969, 2017.

\bibitem{he2016deep}
Kaiming He, Xiangyu Zhang, Shaoqing Ren, and Jian Sun.
\newblock Deep residual learning for image recognition.
\newblock In {\em Proceedings of the IEEE conference on computer vision and
  pattern recognition}, pages 770--778, 2016.

\bibitem{heller2018psBattles}
Silvan Heller, Luca Rossetto, and Heiko Schuldt.
\newblock {The PS-Battles Dataset -- an Image Collection for Image Manipulation
  Detection}.
\newblock {\em CoRR}, abs/1804.04866, 2018.

\bibitem{hinton2015distilling}
Geoffrey Hinton, Oriol Vinyals, and Jeff Dean.
\newblock Distilling the knowledge in a neural network.
\newblock {\em arXiv preprint arXiv:1503.02531}, 2015.

\bibitem{huh2016makes}
Minyoung Huh, Pulkit Agrawal, and Alexei~A Efros.
\newblock What makes imagenet good for transfer learning?
\newblock {\em arXiv preprint arXiv:1608.08614}, 2016.

\bibitem{huh2018fighting}
Minyoung Huh, Andrew Liu, Andrew Owens, and Alexei~A Efros.
\newblock Fighting fake news: Image splice detection via learned
  self-consistency.
\newblock In {\em Proceedings of the European Conference on Computer Vision
  (ECCV)}, pages 101--117, 2018.

\bibitem{jegou2012negative}
Herv{\'e} J{\'e}gou and Ond{\v{r}}ej Chum.
\newblock Negative evidences and co-occurences in image retrieval: The benefit
  of pca and whitening.
\newblock In {\em European conference on computer vision}, pages 774--787.
  Springer, 2012.

\bibitem{jegou2010improving}
Herv{\'e} J{\'e}gou, Matthijs Douze, and Cordelia Schmid.
\newblock Improving bag-of-features for large scale image search.
\newblock {\em International journal of computer vision}, 87(3):316--336, 2010.

\bibitem{jegou2011product}
Herve Jegou, Matthijs Douze, and Cordelia Schmid.
\newblock Product quantization for nearest neighbor search.
\newblock {\em IEEE transactions on pattern analysis and machine intelligence},
  33(1):117--128, 2011.

\bibitem{kingma2014adam}
Diederik~P Kingma and Jimmy Ba.
\newblock Adam: A method for stochastic optimization.
\newblock {\em arXiv preprint arXiv:1412.6980}, 2014.

\bibitem{kornblith2018better}
Simon Kornblith, Jonathon Shlens, and Quoc~V Le.
\newblock Do better imagenet models transfer better?
\newblock {\em arXiv preprint arXiv:1805.08974}, 2018.

\bibitem{KrauseStarkDengFei-Fei_3DRR2013}
Jonathan Krause, Michael Stark, Jia Deng, and Li Fei-Fei.
\newblock 3d object representations for fine-grained categorization.
\newblock In {\em 4th International IEEE Workshop on 3D Representation and
  Recognition (3dRR-13)}, Sydney, Australia, 2013.

\bibitem{OpenImages}
Alina Kuznetsova, Hassan Rom, Neil Alldrin, Jasper Uijlings, Ivan Krasin, Jordi
  Pont-Tuset, Shahab Kamali, Stefan Popov, Matteo Malloci, Tom Duerig, and
  Vittorio Ferrari.
\newblock The open images dataset v4: Unified image classification, object
  detection, and visual relationship detection at scale.
\newblock {\em arXiv:1811.00982}, 2018.

\bibitem{lin2017feature}
Tsung-Yi Lin, Piotr Doll{\'a}r, Ross Girshick, Kaiming He, Bharath Hariharan,
  and Serge Belongie.
\newblock Feature pyramid networks for object detection.
\newblock In {\em Proceedings of the IEEE Conference on Computer Vision and
  Pattern Recognition}, pages 2117--2125, 2017.

\bibitem{lin2014microsoft}
Tsung-Yi Lin, Michael Maire, Serge Belongie, James Hays, Pietro Perona, Deva
  Ramanan, Piotr Doll{\'a}r, and C~Lawrence Zitnick.
\newblock Microsoft coco: Common objects in context.
\newblock In {\em European conference on computer vision}, pages 740--755.
  Springer, 2014.

\bibitem{liu2018deep}
Li Liu, Wanli Ouyang, Xiaogang Wang, Paul Fieguth, Jie Chen, Xinwang Liu, and
  Matti Pietik{\"a}inen.
\newblock Deep learning for generic object detection: A survey.
\newblock {\em arXiv preprint arXiv:1809.02165}, 2018.

\bibitem{lowe2004distinctive}
David~G Lowe.
\newblock Distinctive image features from scale-invariant keypoints.
\newblock {\em International journal of computer vision}, 60(2):91--110, 2004.

\bibitem{mahdian2009using}
Babak Mahdian and Stanislav Saic.
\newblock Using noise inconsistencies for blind image forensics.
\newblock {\em Image and Vision Computing}, 27(10):1497--1503, 2009.

\bibitem{melekhov2018dgc}
Iaroslav Melekhov, Aleksei Tiulpin, Torsten Sattler, Marc Pollefeys, Esa Rahtu,
  and Juho Kannala.
\newblock Dgc-net: Dense geometric correspondence network.
\newblock {\em arXiv preprint arXiv:1810.08393}, 2018.

\bibitem{mishchuk2017working}
Anastasiia Mishchuk, Dmytro Mishkin, Filip Radenovic, and Jiri Matas.
\newblock Working hard to know your neighbor's margins: Local descriptor
  learning loss.
\newblock In {\em Advances in Neural Information Processing Systems}, pages
  4826--4837, 2017.

\bibitem{moreira2018image}
Daniel Moreira, Aparna Bharati, Joel Brogan, Allan Pinto, Michael Parowski,
  Kevin~W Bowyer, Patrick~J Flynn, Anderson Rocha, and Walter~J Scheirer.
\newblock Image provenance analysis at scale.
\newblock {\em IEEE Transactions on Image Processing}, 27(12):6109--6123, 2018.

\bibitem{movshovitz2017no}
Yair Movshovitz-Attias, Alexander Toshev, Thomas~K Leung, Sergey Ioffe, and
  Saurabh Singh.
\newblock No fuss distance metric learning using proxies.
\newblock In {\em Proceedings of the IEEE International Conference on Computer
  Vision}, pages 360--368, 2017.

\bibitem{noh2017large}
Hyeonwoo Noh, Andre Araujo, Jack Sim, Tobias Weyand, and Bohyung Han.
\newblock Large-scale image retrieval with attentive deep local features.
\newblock In {\em Proceedings of the IEEE International Conference on Computer
  Vision}, pages 3456--3465, 2017.

\bibitem{oh2016deep}
Hyun Oh~Song, Yu Xiang, Stefanie Jegelka, and Silvio Savarese.
\newblock Deep metric learning via lifted structured feature embedding.
\newblock In {\em Proceedings of the IEEE Conference on Computer Vision and
  Pattern Recognition}, pages 4004--4012, 2016.

\bibitem{philbin2007object}
James Philbin, Ondrej Chum, Michael Isard, Josef Sivic, and Andrew Zisserman.
\newblock Object retrieval with large vocabularies and fast spatial matching.
\newblock In {\em 2007 IEEE Conference on Computer Vision and Pattern
  Recognition}, pages 1--8. IEEE, 2007.

\bibitem{radenovic2018revisiting}
Filip Radenovi{\'c}, Ahmet Iscen, Giorgos Tolias, Yannis Avrithis, and
  Ond{\v{r}}ej Chum.
\newblock Revisiting oxford and paris: Large-scale image retrieval
  benchmarking.
\newblock In {\em Proceedings of the IEEE Conference on Computer Vision and
  Pattern Recognition}, pages 5706--5715, 2018.

\bibitem{radenovic2016cnn}
Filip Radenovi{\'c}, Giorgos Tolias, and Ond{\v{r}}ej Chum.
\newblock Cnn image retrieval learns from bow: Unsupervised fine-tuning with
  hard examples.
\newblock In {\em European conference on computer vision}, pages 3--20.
  Springer, 2016.

\bibitem{radenovic2018fine}
Filip Radenovi{\'c}, Giorgos Tolias, and Ondrej Chum.
\newblock Fine-tuning cnn image retrieval with no human annotation.
\newblock {\em IEEE transactions on pattern analysis and machine intelligence},
  2018.

\bibitem{razavian2016visual}
Ali~S Razavian, Josephine Sullivan, Stefan Carlsson, and Atsuto Maki.
\newblock Visual instance retrieval with deep convolutional networks.
\newblock {\em ITE Transactions on Media Technology and Applications},
  4(3):251--258, 2016.

\bibitem{redmon2016you}
Joseph Redmon, Santosh Divvala, Ross Girshick, and Ali Farhadi.
\newblock You only look once: Unified, real-time object detection.
\newblock In {\em Proceedings of the IEEE conference on computer vision and
  pattern recognition}, pages 779--788, 2016.

\bibitem{ren2015faster}
Shaoqing Ren, Kaiming He, Ross Girshick, and Jian Sun.
\newblock Faster r-cnn: Towards real-time object detection with region proposal
  networks.
\newblock In {\em Advances in neural information processing systems}, pages
  91--99, 2015.

\bibitem{rocco2017convolutional}
Ignacio Rocco, Relja Arandjelovic, and Josef Sivic.
\newblock Convolutional neural network architecture for geometric matching.
\newblock In {\em Proceedings of the IEEE Conference on Computer Vision and
  Pattern Recognition}, pages 6148--6157, 2017.

\bibitem{rocco2018end}
Ignacio Rocco, Relja Arandjelovi{\'c}, and Josef Sivic.
\newblock End-to-end weakly-supervised semantic alignment.
\newblock In {\em Proceedings of the IEEE Conference on Computer Vision and
  Pattern Recognition}, pages 6917--6925, 2018.

\bibitem{romero2014fitnets}
Adriana Romero, Nicolas Ballas, Samira~Ebrahimi Kahou, Antoine Chassang, Carlo
  Gatta, and Yoshua Bengio.
\newblock Fitnets: Hints for thin deep nets.
\newblock {\em arXiv preprint arXiv:1412.6550}, 2014.

\bibitem{rui1999image}
Yong Rui, Thomas~S Huang, and Shih-Fu Chang.
\newblock Image retrieval: Current techniques, promising directions, and open
  issues.
\newblock {\em Journal of visual communication and image representation},
  10(1):39--62, 1999.

\bibitem{sablayrolles2018spreading}
Alexandre Sablayrolles, Matthijs Douze, Cordelia Schmid, and Herv{\'e}
  J{\'e}gou.
\newblock Spreading vectors for similarity search.
\newblock {\em arXiv preprint arXiv:1806.03198}, 2018.

\bibitem{salloum2018image}
Ronald Salloum, Yuzhuo Ren, and C-C~Jay Kuo.
\newblock Image splicing localization using a multi-task fully convolutional
  network (mfcn).
\newblock {\em Journal of Visual Communication and Image Representation},
  51:201--209, 2018.

\bibitem{sharif2014cnn}
Ali Sharif~Razavian, Hossein Azizpour, Josephine Sullivan, and Stefan Carlsson.
\newblock Cnn features off-the-shelf: an astounding baseline for recognition.
\newblock In {\em Proceedings of the IEEE conference on computer vision and
  pattern recognition workshops}, pages 806--813, 2014.

\bibitem{singh2018sniper}
Bharat Singh, Mahyar Najibi, and Larry~S Davis.
\newblock Sniper: Efficient multi-scale training.
\newblock In {\em Advances in Neural Information Processing Systems}, pages
  9333--9343, 2018.

\bibitem{sivic2003video}
Josef Sivic and Andrew Zisserman.
\newblock Video google: A text retrieval approach to object matching in videos.
\newblock In {\em null}, page 1470. IEEE, 2003.

\bibitem{teichmann2018detect}
Marvin Teichmann, Andre Araujo, Menglong Zhu, and Jack Sim.
\newblock Detect-to-retrieve: Efficient regional aggregation for image search.
\newblock {\em arXiv preprint arXiv:1812.01584}, 2018.

\bibitem{tolias2015particular}
Giorgos Tolias, Ronan Sicre, and Herv{\'e} J{\'e}gou.
\newblock Particular object retrieval with integral max-pooling of cnn
  activations.
\newblock {\em arXiv preprint arXiv:1511.05879}, 2015.

\bibitem{wang2017deep}
Jian Wang, Feng Zhou, Shilei Wen, Xiao Liu, and Yuanqing Lin.
\newblock Deep metric learning with angular loss.
\newblock In {\em Proceedings of the IEEE International Conference on Computer
  Vision}, pages 2593--2601, 2017.

\bibitem{WelinderEtal2010}
P. Welinder, S. Branson, T. Mita, C. Wah, F. Schroff, S. Belongie, and P.
  Perona.
\newblock {Caltech-UCSD Birds 200}.
\newblock Technical Report CNS-TR-2010-001, California Institute of Technology,
  2010.

\bibitem{wen2016coverage}
Bihan Wen, Ye Zhu, Ramanathan Subramanian, Tian-Tsong Ng, Xuanjing Shen, and
  Stefan Winkler.
\newblock Coverage — a novel database for copy-move forgery detection.
\newblock In {\em 2016 IEEE International Conference on Image Processing
  (ICIP)}, pages 161--165. IEEE, 2016.

\bibitem{wu2017sampling}
Chao-Yuan Wu, R Manmatha, Alexander~J Smola, and Philipp Krahenbuhl.
\newblock Sampling matters in deep embedding learning.
\newblock In {\em Proceedings of the IEEE International Conference on Computer
  Vision}, pages 2840--2848, 2017.

\bibitem{wu2018busternet}
Yue Wu, Wael Abd-Almageed, and Prem Natarajan.
\newblock Busternet: Detecting copy-move image forgery with source/target
  localization.
\newblock In {\em Proceedings of the European Conference on Computer Vision
  (ECCV)}, pages 168--184, 2018.

\bibitem{ye2007detecting}
Shuiming Ye, Qibin Sun, and Ee-Chien Chang.
\newblock Detecting digital image forgeries by measuring inconsistencies of
  blocking artifact.
\newblock In {\em 2007 IEEE International Conference on Multimedia and Expo},
  pages 12--15. IEEE, 2007.

\bibitem{yosinski2014transferable}
Jason Yosinski, Jeff Clune, Yoshua Bengio, and Hod Lipson.
\newblock How transferable are features in deep neural networks?
\newblock In {\em Advances in neural information processing systems}, pages
  3320--3328, 2014.

\bibitem{yue2015exploiting}
Joe Yue-Hei~Ng, Fan Yang, and Larry~S Davis.
\newblock Exploiting local features from deep networks for image retrieval.
\newblock In {\em Proceedings of the IEEE conference on computer vision and
  pattern recognition workshops}, pages 53--61, 2015.

\bibitem{zhang2017learning}
Xu Zhang, Felix~X Yu, Sanjiv Kumar, and Shih-Fu Chang.
\newblock Learning spread-out local feature descriptors.
\newblock In {\em Proceedings of the IEEE International Conference on Computer
  Vision}, pages 4595--4603, 2017.

\bibitem{zheng2018sift}
Liang Zheng, Yi Yang, and Qi Tian.
\newblock Sift meets cnn: A decade survey of instance retrieval.
\newblock {\em IEEE transactions on pattern analysis and machine intelligence},
  40(5):1224--1244, 2018.

\bibitem{zhou2018generate}
Peng Zhou, Bor-Chun Chen, Xintong Han, Mahyar Najibi, and Larry~S Davis.
\newblock Generate, segment and replace: Towards generic manipulation
  segmentation.
\newblock In {\em Proceedings of the AAAI conference on artificial
  intelligence}, 2020.

\bibitem{zhou2018learning}
Peng Zhou, Xintong Han, Vlad~I Morariu, and Larry~S Davis.
\newblock Learning rich features for image manipulation detection.
\newblock In {\em Proceedings of the IEEE Conference on Computer Vision and
  Pattern Recognition}, pages 1053--1061, 2018.

\end{thebibliography}
}

\end{document}